\documentclass{evolang}
\usepackage{graphicx}
\usepackage{url}
\usepackage{authblk}
\begin{document}

\title{Failures and Successes to Learn a Core Conceptual Distinction from the Statistics of Language\setcounter{footnote}{1}\thanks{\uppercase{P}ublished at the 15th \uppercase{I}nternational \uppercase{C}onference on the \uppercase{E}volution of \uppercase{L}anguage (\uppercase{E}volang \uppercase{XV}).}}
\author[1]{Zhimin Hu}
\author[1]{Jeroen van Paridon}
\author[*1]{Gary Lupyan}
\affil[*]{corresponding author}
\affil[1]{Department of Psychology, University of Wisconsin-Madison, Madison, USA}

\maketitle

\begin{abstract}
Generic statements like “tigers are striped” and “cars have radios” communicate information that is, in general, true. However, while the first statement is true *in principle*, the second is true only statistically. People are exquisitely sensitive to this principled-vs-statistical distinction. It has been argued that this ability to distinguish between something being true by virtue of it being a category member versus being true because of mere statistical regularity, is a general property of people’s conceptual machinery and cannot itself be learned. We investigate whether the distinction between principled and statistical properties can be learned from language itself. If so, it raises the possibility that language experience can bootstrap core conceptual distinctions and that it is possible to learn sophisticated causal models directly from language. We find that language models are all sensitive to statistical prevalence, but struggle with representing the principled-vs-statistical distinction controlling for prevalence. Until GPT-4, which succeeds.

\end{abstract}

\section{Introduction}
People interpret generic statements such as \textit{airplanes have wings} and \textit{dogs bark} to mean that the named property is, in general, true of the category \shortcite{hollander_generic_2009}. Other statements of this form, however, such as \textit{airplanes carry passengers} and \textit{dogs wear collars}, while also being judged as generally true, have a decidedly different quality. In a series of papers, Prasada and colleagues \shortcite{prasada2006principled,prasada_mechanisms_2016,prasada2013conceptual} drew a distinction between generics that express \textit{principled} properties and generics that express merely \textit{statistical} properties. A statement expressing a principled property, such as \textit{airplanes have wings}, retains its truthfulness when asked whether it is true because of (or by virtue of) being that thing. For example, in the experiments we describe below, on a scale of -3 = completely false to +3 = completely true, people judged the statement \textit{airplanes have wings} with a mean of 2.9. This declines only slightly if asked whether it is true that airplanes have wings \textit{because they are airplanes} (M=2.6). A statement like \textit{airplanes have passengers} is judged to also be mostly true (M=1.8), but if asked whether airplanes have passengers \textit{because} they are airplanes, the truth estimate drops (M=0.6). Importantly, this key result remains when one controls for confounds such as prevalence and cue-validity, showing that it is not simply an artifact of principled connections being more common or it being harder to come up with counter-examples.

Results like these have been used to argue that people's ability to distinguish between principled and statistical generics requires an \textit{a priori} sensitivity to a distinction between statistical vs. ``in-principle'' properties. Because there are no structural differences between generics that could inform this distinction, it is thought that the distinction cannot be learned through associations \cite<see>{prasada2013conceptual,haward2018development}, and perhaps cannot even be represented by an associative mechanism \cite{prasada_physical_2021}.

However, even though generic statements do not encode the principled/statistical distinction in their structure, the distinction might still be captured in the distributional structure of language itself. In this study, we investigated whether the statistical/generic distinction is recoverable from the statistics of language. We did this by predicting human judgments of generic statements from judgments derived from distributional language models. Finding that this distinction can be learned by an associative mechanism from language alone is important for two main reasons. First, it shows that it is \textit{in principle} possible to learn a formal conceptual distinction argued to be unlearnable (and even unrepresentable) by an associative mechanism. Second, it opens the door to asking questions of key interest to the study of language evolution: (1) Are languages structured to facilitate extracting principled item-property relationships? (2) Where in language is such information represented? (3) Are languages not only a \textit{source} of generic information \cite{rhodes_cultural_2012}, but do they also help structure the very core of our conceptual system?

To anticipate our results, we find that language models are all sensitive to item prevalence. Statements probing frequent item-property combinations like \textit{oranges grow on trees} and \textit{kangaroos have pouches} are judged by models as more true than statements probing rarer item-property combinations such as \textit{professors are absent-minded} and \textit{birds are kept in cages}. However, a distinction in truth judgments between principled and statistical relations, when controlling for prevalence and cue-validity, only appeared for the largest language models we tested. 

\section{Human ratings}
We began by constructing a corpus of 208 generic statements and having them rated on several scales using a procedure adapted from \citeNP{prasada2013conceptual}.

\subsection{Participants}
We recruited 91 native speakers of English residing in the United States through Amazon Mechanical Turk in exchange for a \$2 payment. Seven participants were rejected for failing basic attention checks, leaving 84 participants.

\subsection{Procedure}
Participants were asked to judge four different aspects of generic statements, sentences describing properties of objects, people, and animals: (1) \textit{Bare generic truth judgment:} ``How true is the following statement: \emph{Airplanes have seatbelts.}''; (2) \textit{By-virtue-of truth judgment}: ``How true is the following statement: \emph{Because they are airplanes, airplanes have seatbelts.}''; (3)
\textit{Prevalence rating}: ``Think of airplanes, how likely are they to have seatbelts?''; (4): \textit{Cue validity rating}: ``You learn that [unknown things/people] have wings, how likely is it [are they] to be airplanes?'' Each participant was presented with 26 statements of each type. The statements were counterbalanced across participants and rating questions so that no participant rated a given generic more than once.

\begin{figure}[ht]
\begin{center}
\fbox{\includegraphics[width=\columnwidth]{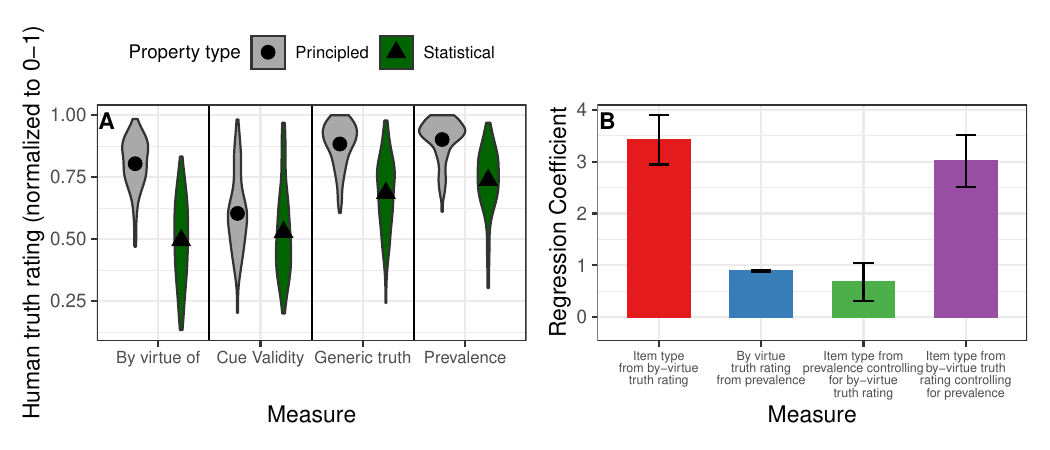}}
\caption{A. Mean human truth ratings for each sentence frame, comparing principled and statistical relationships. B. Regression coefficients (with SEs) showing key relationships between truth ratings, property type, and prevalence (see text.)}
\label{combined_plot}
\end{center}
\end{figure}
\vspace{-15pt}

\subsection{Results}
Our results, shown in Fig. \ref{combined_plot}, closely replicate the findings of \citeNP{prasada2013conceptual}. Bare generics (``Airplanes have wings``) expressing principled relationships are rated as more true than statements expressing statistical relationships (``Airplanes have passengers'') and the same goes for by-virtue-of judgments (but more so). By-virtue truth ratings \textit{are} affected by prevalence (the blue bar in Fig. \ref{combined_plot}B), and prevalence predicts item-type (principled vs. statistical) when controlling for the by-truth rating (green bar). Importantly, the ability of by-virtue judgments to predict property type (red bar) is nearly undiminished when we control for prevalence (cf. red and purple bars). We will be comparing the model results to this U-shaped pattern of coefficients shown in Fig. \ref{combined_plot}B.

\section{Can the principled/statistical distinction be learned from language itself?}
To determine whether distributional models can differentiate between statistical and principled generics, we predicted property type (statistical vs. principled) from the cosine similarity between the target word and the property. 

\subsection{Models}
We tested the language models listed Table~\ref{table1} using the Huggingface implementations of BERT \shortcite{devlin2018bert}, ALBERT \shortcite{lan2019albert}, DistilBERT \shortcite{sanh2019distilbert}, RoBERTa \shortcite{liu2019roberta}, GPT \shortcite{radford2018improving}, and GPT-2 \shortcite{radford2019language}. We used the OpenAI APIs for GPT-3.5 and GPT-4. 
\begin{table}[h]
\tablecaption{Overview of the models we tested}
{
\begin{tabular}{@{}cccc@{}}
\hline
Model name  & Training sources & Size of training corpus & \# Number of parameters \\
\hline
BERT (base) & Wiki, books & 3.3B tokens (13 GB data) & 110M \\
ALBERT (base-v1) & Wiki, books & 3.3B tokens (13 GB data) & 11M \\
Distilbert (base)  & Wiki, books & 3.3B tokens (13 GB data) & 66M \\
RoBERTa (base) & Wiki, books, web crawl & 161 GB data & 125M \\
GPT & Web crawl & 800M tokens & 110M \\
GPT-2 (base) & Web crawl, Reddit, & 8M documents (40 GB data) & 117M \\
GPT-3.5 & Unknown superset of GPT-2  & Unknown & Unknown \\
GPT-4 & Unknown superset of GPT-3.5 & Unknown & Unknown \\
\hline
\end{tabular}

\label{table1}}
\end{table}

\subsection{Methods}
To measure the represented similarity between the target words and their properties, we first needed to obtain their model embeddings. Because the transformer models only generate contextual embeddings, we simulated a decontextualized context by using the "all but the top" method proposed by \shortcite{mu2018all}. This method removes the top k principal components (here, k=7) as computed by sampling additional corpuses of text from the NLI dataset \shortcite{bowman-etal-2015-large} and wiki-103 \shortcite{merity2016pointer}. It ensures the resulting embeddings reflect a more contrastive meaning of a given phrase. The models' truth judgment was then operationalized as the cosine similarity between the target-word (e.g., ''airplanes'') and the property (''have wings''). 

Because GPT-3.5 and 4 are fine-tuned for question-answering, it was possible to probe their `knowledge' more directly by having them rate the generics using the same prompt as human participants. The models received the following type of prompt: \textit{return only one integer between -3 and 3, where -3 means the sentence is definitely false and 3 means the sentence is definitely true: Because they are airplanes, airplanes have wings}. We tested each of the 208 generics 15 times and averaged the ratings. The variance of this average was less than 0.01.

\begin{figure}[ht]
\begin{center}
\fbox{\includegraphics[width=.7\linewidth]{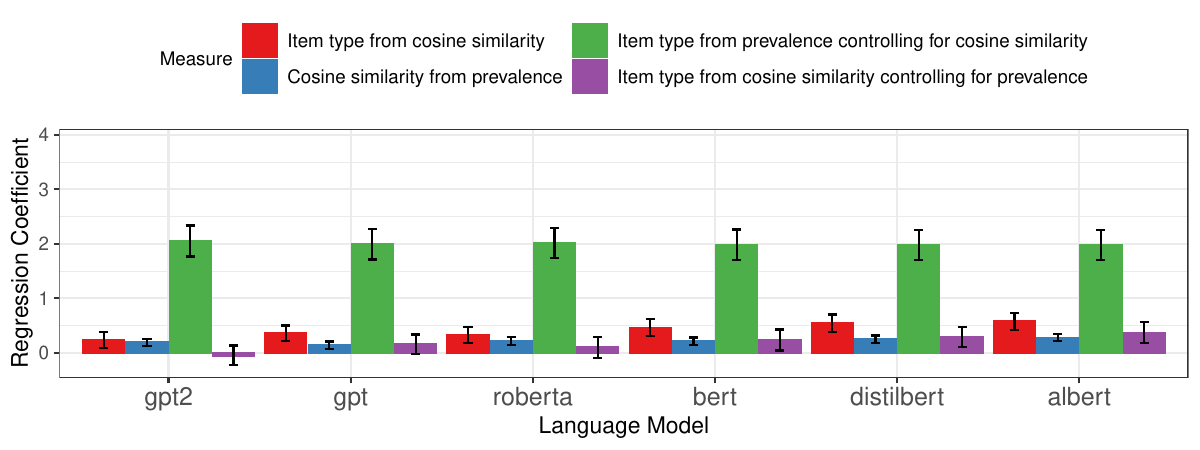}}
\caption{Regression coefficients (with SEs) indicating relationships between item-property cosine-similarity, property-type, and prevalence using the analogous models used in Fig. 2.}
\label{cosines}
\end{center}
\end{figure}
\vspace{-15pt}

\begin{figure}[ht]
\begin{center}
    \fbox{\includegraphics[width=0.55\linewidth]{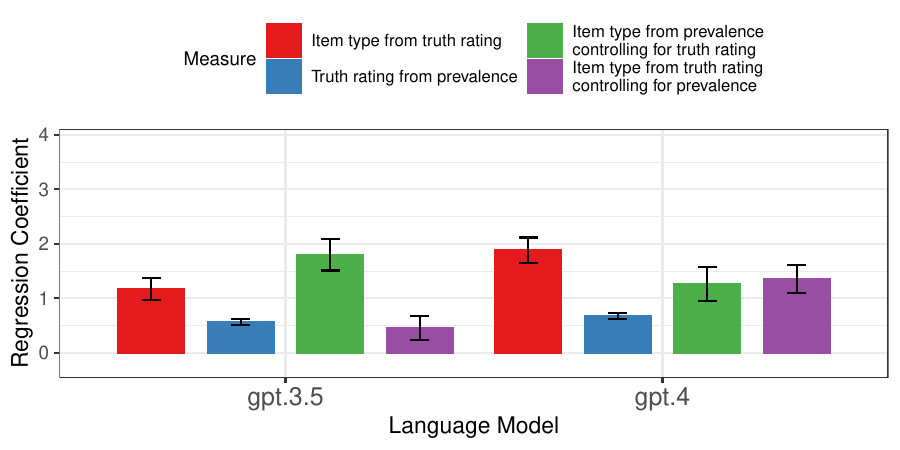}}
    \caption{Regression coefficients (with SEs) indicating relationships between model-generated by-virtue-of truth ratings, property-type and human-ratings of prevalence.}
    \label{gpt3.5_and_4}
\end{center}
\end{figure}
\vspace{-15pt}

\section{Results}
The basic pattern of results from the cosine similarity analyses is shown in Fig. \ref{cosines}. Across all six models, we see the same qualitative pattern. The models distinguish between principled and statistical connections: the similarity between the target word like `airplane' and a principled property like `wings' is greater than a statistical property like `passengers (red bars). However, when we control for prevalence, this association largely disappears (purple bars; it is only marginally above 0 in ALBERT). Human truth ratings (especially by-virtue ratings) are much better predictors of property type than the prevalence ratings. For the models, this is not the case, as indicated by the large green bar in comparison to \ref{combined_plot}B.

Turning to our experiments with GPT-3.5 and GPT-4 in which we were able to directly query their truth judgments, we find a rather different result (Fig. \ref{gpt3.5_and_4}). In GPT-3.5, truth ratings are barely predictive of property-type when controlling for prevalence (green bar; \textit{t}=2.05, \textit{p}=.04), while for GPT-4 they remain strongly predictive, \textit{t}=5.18, \textit{p} \textless.00001). As a complementary analysis, we examined by-item relationships. For each item (e.g., dogs, trampolines, trumpets), we can compare the by-human virtue-of truth judgment for the principled vs. statistical statement, and compare it to the cosine-similarity-based measure for the BERT-type models and to the truth judgments for the GPT models. We find correlations ranging from .21 for BERT to .28 for DistilBERT. These increase to .49 for GPT-3.5 and to .61 for GPT-4.

\section{General Discussion}
People know that airplanes have wings and carry passengers, and simultaneously know that the former but not the latter property is part of what \textit{it means} to be an airplane. Since this distinction is not marked in language, it has been thought that it must come from elsewhere, such as an innate generative type-token mechanism \cite{prasada_mechanisms_2016}. We show here that it is, in principle, possible to learn this distinction from the statistics of language, but it is far from trivial, emerging most clearly only in GPT-4. All tested transformer models trained on English text were sensitive to prevalence, as shown by significant associations between prevalence and cosine similarity/model truth judgments. People's judgments also show sensitivity to prevalence, which makes sense since it is often a good proxy for whether a relationship is principled or statistical: that \textit{principled} relationships have, on average, considerably higher prevalence than merely \textit{statistical} ones. But human judgments continue to strongly distinguish principled and statistical relationships when prevalence is partialled out--consistent with the view that people base their judgments on causal models, presumably learned from rich multimodal experience \cite<see e.g.>{prasada2006principled,prasada2013conceptual}. The failure of language models to distinguish statistical from principled properties once prevalence is partialled out indicates that the models are basing their 'judgments' on statistical co-occurrence. And yet, when we test more recent models such as GPT-3.5 and especially GPT-4, the picture starts to shift, consistent with the possibility that lowering next-token prediction error at scale can lead to the models inducing more sophisticated world models \shortcite<e.g.,>{li2022emergent,mirchandani2023large,michaelov2023can,li2021implicit}. Although it is unknown at present what allows GPT-4 to succeed, our experiments provide an in-principle proof that it is possible to induce sophisticated causal models of item-property relations from language alone. Although it is rather unlikely that people learn the distinction between principled and statistical properties from language in the same way, our results hint that input from language may be more instrumental for laying down core conceptual distinctions than previously thought.

\bibliographystyle{apacite}
\bibliography{evolang} 

\end{document}